\documentclass{article}

\usepackage{graphicx}
\usepackage{epstopdf}
\usepackage{amssymb}
\usepackage{algorithmic,algorithm}

\usepackage{natbib}
\begin{document}

\title{Event management for large scale event-driven digital hardware spiking neural networks}
\author{Louis-Charles~Caron \and Michiel~D'Haene \and Fr\'ed\'eric~Mailhot \and Benjamin~Schrauwen \and Jean~Rouat}
\date{\today}
\maketitle

\begin{abstract}
The interest in brain-like computation has led to the design of a plethora of innovative neuromorphic systems. Individually, spiking neural networks (SNNs), event-driven simulation and digital hardware neuromorphic systems get a lot of attention. Despite the popularity of event-driven SNNs in software, very few digital hardware architectures are found. This is because existing hardware solutions for event management scale badly with the number of events. This paper introduces the structured heap queue, a pipelined digital hardware data structure, and demonstrates its suitability for event management. The structured heap queue scales gracefully with the number of events, allowing the efficient implementation of large scale digital hardware event-driven SNNs. The scaling is linear for memory, logarithmic for logic resources and constant for processing time. The use of the structured heap queue is demonstrated on field-programmable gate array (FPGA) with an image segmentation experiment and a SNN of 65~536 neurons and 513~184 synapses. Events can be processed at the rate of 1 every 7 clock cycles and a 406$\times$158 pixel image is segmented in 200 ms.
\end{abstract}

\section{Introduction} \label{sec:Intro}
Neuromorphic systems attempt to mimic the way the brain processes information and are the test benches of new theories of the neural paradigm, exploration tools for connectionists and very intriguing computation platforms. Neuromorphic systems can take many forms. They have been implemented on graphical processing units (GPU) \citep{NageswaranEtAl2009}, digital signal processors (DSP) \citep{PlanaEtAl2007}, analog very-large-scale integration (aVLSI) circuits \citep{BasuEtAl2010}, digital VLSI (dVLSI) circuits \citep{SeoEtAl2011}, field-programmable gate arrays (FPGA) \citep{SchrauwenEtAl2008,ThomasLuk2009,CaronEtAl2011}, mixed-signal VLSI circuits \citep{CawleyEtAl2011,MoradiIndiveri2011} and as software-hardware coprocessors \citep{RosEtAl2006}. Ultimately, the choice of the implementation platform depends on the application and the designer's specific needs. This paper focuses on digital hardware (dVLSI circuits and FPGAs) and aims at laying the ground for the design of large-scale embedded event-driven spiking neural networks.

Spiking neurons, the third generation of artificial neurons, are more computationally powerful than the previous generations \citep{Maass1997}. A spiking neural network (SNN) is a collection of dynamical systems, the neurons, that affect each other through point-to-point links called synaptic connections. Spiking neurons are characterized by one or several state variables. The state of a spiking neuron changes over time and when it meets a certain condition, it emits a spike. Spikes are delivered to other neurons through the synaptic connections and are scaled by a factor, the synaptic weight. The neuron model specifies the equations that govern the evolution of the neuron's state variables in time, the condition for emitting a spike and the effect of receiving and emitting a spike. The network topology defines the interconnection pattern of the neurons and the value of the synaptic weights.

\algsetup{indent=2em}
\begin{algorithm}[h!]
\caption{Time-driven implementation}\label{alg:TD}
\begin{algorithmic}[1]
\medskip
\REPEAT
\STATE move one step forward in time\label{algline:TDstep}
\FOR {each neuron in the network}\label{algline:TDfor}
\STATE update state\label{algline:TDtime}
\ENDFOR
\UNTIL {desired time is reached}
\end{algorithmic}
\end{algorithm}
\algsetup{indent=2em}
\begin{algorithm}[h!]
\caption{Event-driven implementation}\label{alg:ED}
\begin{algorithmic}[1]
\medskip
\REPEAT
\STATE move one event forward in time\label{algline:EDstep}
\FOR {each neuron involved in the event}\label{algline:EDfor}
\STATE update state\label{algline:EDtime}
\ENDFOR
\STATE find next event to happen\label{algline:EDqueue}
\UNTIL {desired time is reached}
\end{algorithmic}
\end{algorithm}
A SNN implementation is the specific way in which time, the neuron model and spikes, or events, are handled in order to calculate the network's state at a desired point in time. Two different implementations are widely used and opposed: time-driven and event-driven. Both strategies are described in algorithm \ref{alg:TD} and algorithm \ref{alg:ED}. A fundamental difference between both strategies lies in the time increment, at line \ref{algline:TDstep}. In a time-driven SNN, time is increased by a constant amount. The update of the neurons' state at line \ref{algline:TDtime} is straightforward, but the events occurring during a given time step are processed as if they occurred at the same point in time, which might not be exact. The choice of the time step is a compromise between processing speed and temporal accuracy. In some cases, it is possible to use temporal interpolation and restore a more accurate time of occurrence of the events \citep{MorrissonEtAl2007}. In the event-driven implementation, the time steps fit the occurrence of events and optimal precision in time can be achieved. The state update of line \ref{algline:EDtime} is more mathematically involved as the step size is variable. Also, progression in time can be slow if the average number of events occurring in the network per unit of time is high. A second difference is the number of neurons processed at each iteration of the for loop of line \ref{algline:TDfor}. In the time-driven algorithm, each neuron in the network is updated to the current time step. The event-driven strategy avoids this costly update of the whole population of neurons. It exploits the fact that if a neuron does not receive or emit a spike, then its state is defined by the dynamical equations of the neuron model. In an event-driven SNN, the state of a neuron is represented by the time at which it should spike next, as determined by the neuron model and without consideration for possible interactions with other neurons. This value, the predicted firing time, doesn’t change as time passes, but only when a neuron is involved in an event. During an iteration of the processing loop, only the neurons affected by the current event are processed. Simple neuron models can be inverted in order to calculate the predicted firing times \citep{MorrissonEtAl2005}. With more complex neuron models, it might be necessary to use iterative procedures \citep{DHaeneEtAl2009} or to approximate the time of occurrence of events. Lastly, the event-driven algorithm involves one extra operation: identifying the next event to happen (line \ref{algline:EDqueue}). At every iteration of the main loop, an event is processed: it is removed from the list of predicted events and the state of the involved neurons is modified. As a result, some predicted events get delayed, anticipated or cancelled, and new events are created. The event queue is the functional block whose role is to keep the list of events up to date and identify which one is the next to happen. It is a hybrid memory and sorting algorithm. In an event-driven SNN, the event queue has to carry the following operations: output the next event to happen, allow the insertion of new events, allow the deletion of existing events and allow the modification of existing events. The last operation, called an update, is optional since it can be replaced by the deletion of the event to modify followed by the insertion of the same event with the updated information. The main focus of this paper is to describe an efficient way to implement the event queue in a digital hardware event-driven SNN.

The literature on software SNNs abounds in event-driven systems \citep{Pratt1990,Watts1994,GrassmannAnlauf1999,MattiaDelgiudice2000,LeeFarhat2001,DelormeThorpe2003,MouraudPuzenat2009}. In addition to implementing different neuron models and network topologies, each software SNN has its own way of managing the event queue. \citet{Pratt1990} and \citet{Watts1994} use a sorted list to implement it. \citet{GrassmannAnlauf1999} do not specify how the event queue is managed. \citet{MattiaDelgiudice2000} use an array of FIFOs, \citet{LeeFarhat2001} use two heap queues, \citet{DelormeThorpe2003} use a pseudo-sorting algorithm by regular sampling and \citet{MouraudPuzenat2009} use a variation of the calendar queue. On the hardware side, most architectures rely on a fixed time step approach to compute the neural dynamics and do not predict future events \citep{Bako2009,CawleyEtAl2011,CassidyEtAl2011,SeoEtAl2011,CheungEtAl2012}. \citet{Mehrtash2003} and \citet{Schoenauer2002} use a variable step size for neuron state update, but only predict spikes occurring during the next time step. To our knowledge, only \citet{AgisEtAl2007} follow a pattern similar to algorithm \ref{alg:ED} and truly benefits from the advantages of the event-driven strategy in a digital hardware SNN.

The fact that existing software event-driven SNNs use such a wide variety of algorithms to implement the event queue suggests that it plays a very important role and must be carefully designed. We postulate that the reason why so few hardware event-driven SNNs exist is because no digital hardware algorithm can efficiently fill the role of the event queue. The solution chosen by \citet{AgisEtAl2007}, is to use an unsorted list, i.e. to store each event in memory to a designated place. By accessing the right address in this memory, any event can be read, inserted, deleted and modified. The identification of the next event to happen is done on demand by scanning the whole event queue and using a pipelined comparator tree to find the one with smallest time of occurrence. This strategy results in a $O(N)$ complexity in memory and $O(N)$ complexity in logic, where $N$ is the capacity of the queue (maximum number of events). Read, insert, delete and update operations have a $O(1)$ complexity in time, and the search for the next event to happen has $O(\log(N))$ complexity in time. That is, as the capacity of the event queue is increased, it takes more and more clock cycles to find the smallest time of occurrence in the list of events.

This performance degradation as the network size increases can be avoided. In this paper, we introduce the Structured Heap Queue (SHQ) data structure as a candidate to implement the event queue in a large scale digital hardware event-driven SNN. The SHQ has a $O(N)$ complexity in memory, $O(\log(N))$ complexity in logic and $O(1)$ complexity in time. The $O(1)$ complexity in time holds for all important operations on the queue and means that the number of clock cycles required for managing the event queue is independent of its size. We demonstrate the use of the SHQ in a FPGA SNN implementation and an image segmentation task. We show the result of a pixel-based image segmentation realized with the described system to illustrate the full design process of a digital event-driven SNN using the SHQ, from the neuron model to the application. The SHQ is an event management algorithm and can be coupled to any digital event-driven SNN. It is not restricted to the SNN implementation described in this paper.

In section \ref{sec:SHQ}, the SHQ data structure is introduced. The design of the queue, the operations it supports, and the way to pipeline them in a digital hardware implementation are described. A memory-optimized version of the SHQ is also detailed in section \ref{sec:MemOpt}. The SNN we use to demonstrate the SHQ is presented in section \ref{sec:SNN}. The hardware implementation of the SNN is described in section \ref{sec:Implementation}. The performance and resource usage of the system are presented in section \ref{sec:EDNS_Results}. Section \ref{sec:Discussion} discusses the performance of the SHQ and its use as an event queue and section \ref{sec:Conclusion} concludes the paper.

\section{The Structured Heap Queue}\label{sec:SHQ}
The structured heap queue (SHQ) is an implicit data structure derived from the binary heap queue \citep{Gonnet1975}. Because we focus on hardware SNNs, we describe and evaluate the SHQ in this section by comparing it to the pipelined heap queue (PHQ) \citep{IoannouKatevenis2007}, a hardware implementation of the heap queue (see also \citep{BhagwanLin2000}). In a PHQ, a set of elements is partially sorted in order to find the one with lowest (or highest) value. To facilitate the link with the event queue of an event-driven SNN, assume an element consists of two fields: a unique identification number (ID) and a value. In the explanations of section \ref{sec:EventQueue}, an element's ID is a neuron number and its value is the neuron's predicted firing time. A PHQ takes the form of a binary memory tree with one 2-input comparator per level of the tree and some extra logic and registers for control. The PHQ supports two operations: insert a new element and delete the element in the root node. These operations are designed such that their execution maintains the heap property, that is, the guarantee that any node in the tree contains an element of smaller value than that of its 2 children nodes. The operations also keep the binary tree balanced, meaning that the elements spread in the tree over as few levels as possible. These two properties ensure that reading the root node will yield the element with lowest value, and that the tree is as compact as possible. The PHQ has $O(N)$ complexity in memory and $O(\log(N))$ complexity in logic, with $N$ the number of elements. It supports three operations, insert, delete element with lowest value and read element with lowest value, which all have $O(1)$ complexity in time. To use the PHQ as an event queue, it must be possible to delete an element with a given ID from the tree, even if it is not the element with lowest value. In the PHQ, one option is to scan the entire memory tree to find the element with corresponding ID in order to delete it, an operation with $O(N)$ complexity in time.

The capability of finding arbitrary elements in the tree is required in all important operations of an event queue. The linear complexity in time of this operation in a PHQ would lead to catastrophic performance in a large scale SNN. However, the compactness of the tree, i.e. minimal memory usage, is not of the utmost importance for an event queue. The balanced property of the PHQ can be dropped to gain a degree of freedom in the placement of the elements in the tree. In a SHQ, this extra degree of freedom is used such that the search for an arbitrary element becomes a $O(1)$ complexity in time operation.

\begin{figure}[ht!]
\centering
\includegraphics{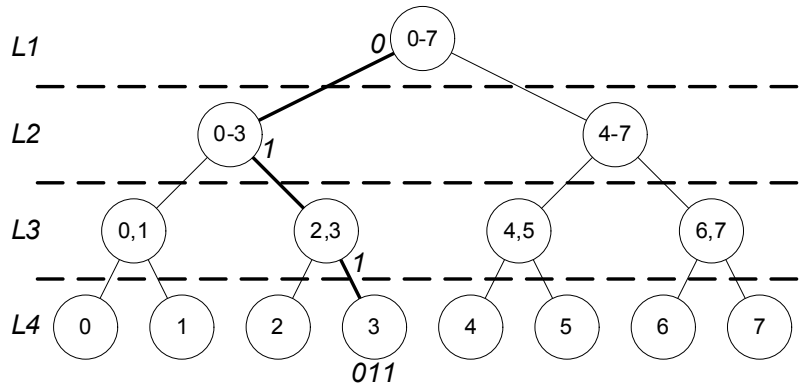}
\caption[Schematic representation of the binary memory tree in a structured heap queue]{Schematic representation of the binary memory tree in a structured heap queue. Each element moves down the tree following a unique path from the root node to one of the leaf nodes. The path an element can follow is determined by its identification number (ID). The numbers in a node indicate which elements can possibly occupy this node. Any element can be stored in the root node, but each leaf node is specific to only one element ID. The path associated with an element is found by branching either left or right depending on the binary representation of the element's ID, starting with the left-most bit when branching from the root node. The path associated to element \#3 is shown in bold.}
\label{fig:SchemNaiveSHQ}
\end{figure}

The SHQ is an unbalanced PHQ in which any given element is constrained to be placed in a specific part of the tree, determined by the element's ID. The region of the tree assigned to an element is called a path, as it starts at the root node and ends at one of the leaf nodes. An element's path is found by scanning its ID from left to right, and progressing down the tree by branching either left or right depending on the value of the individual bits. This defines a unique path for each element and, when searching for an element with a given ID, one single node per level has to be checked. Figure \ref{fig:SchemNaiveSHQ} shows a schematic representation of the binary memory tree and highlights the path of element \#3. This modification comes with two disadvantages. Operations take more clock cycles to execute in the SHQ, but the complexity in time remains $O(1)$. The SHQ has higher memory requirements than the PHQ, but maintains a $O(N)$ complexity in memory.

\subsection{Operations in the Structured Heap Queue}\label{sec:Mechanics}
The SHQ supports 3 operations: insert new element, delete element with given ID and read element with given ID. An update operation, used to change the value of an existing element in the queue, can be executed by issuing a delete followed by an insert operation. All operations in a structured heap queue start at the root node and progress down the tree, one level at a time. An operation follows the path of the element it is executed on and, on each level of the tree, either promotes (move up) or demotes (move down) one element in order to maintain the heap property. Promoting and demoting elements require three steps: reading the content of one or two nodes, comparing two elements, and writing back the appropriate values to memory. A description of the delete, read and insert operations follows.

\subsubsection{Delete and read operations}
\begin{figure}[ht!]
\centering
\includegraphics[width=6cm]{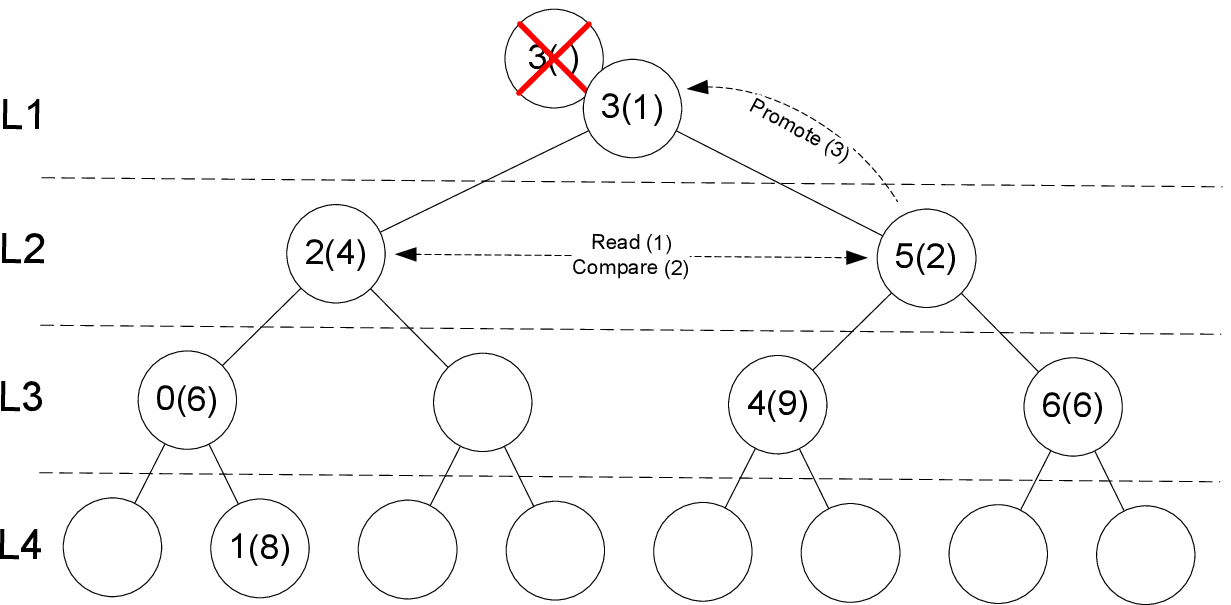}
\includegraphics[width=6cm]{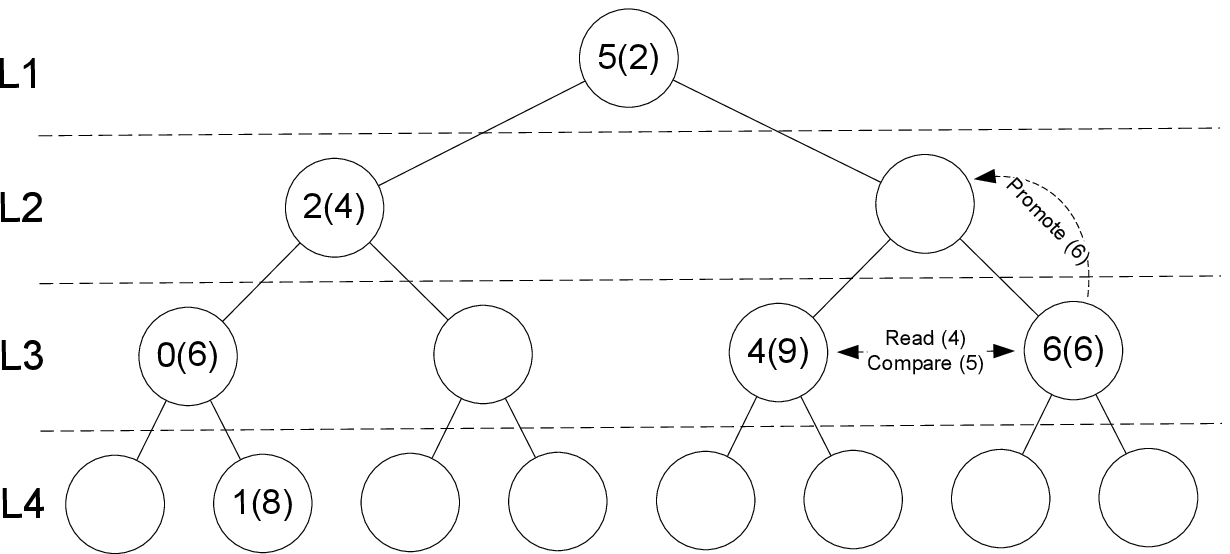}
\includegraphics[width=6cm]{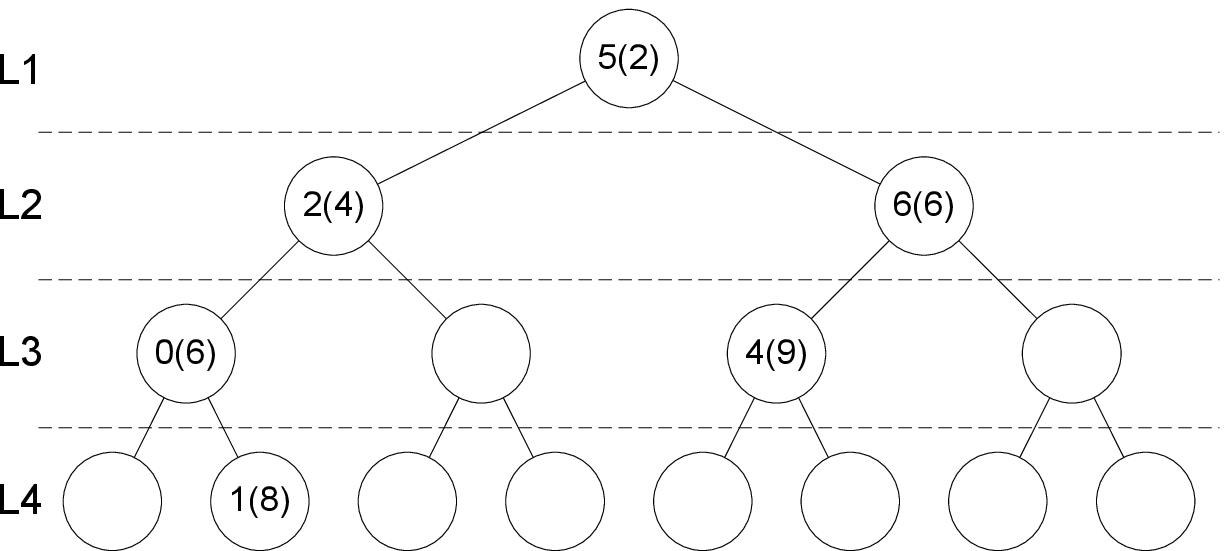}
\caption[Delete operation in the structured heap queue]{Delete operation in the structured heap queue. The pair of numbers in a node indicates the identification number (ID) and, inside parentheses, the value of the element. The operation consists in the deletion of the element \#3. The operation's steps are indicated on the arrows along with their order in parentheses. \textbf{Top left} Initial state of the queue and first steps of the delete operation. Element \#3 is in the root node and thus the locate phase does not take place. The nodes containing elements \#2 and \#5 are read and their value is compared. Element \#5 is promoted to replace element \#3. \textbf{Top right} Last steps of the delete operation. The nodes containing elements \#4 and \#6 are read and their value is compared. Element \#6 is promoted. \textbf{Bottom} Final state of the queue after the delete operation.}
\label{fig:DeleteNaiveSHQ}
\end{figure}

The delete operation is divided into two phases: locate and promote. During the locate phase, only a read step needs to be executed on each level as the operation progresses down the tree. Once the node containing the element to delete is found, the promote phase starts. The two elements in the children nodes of the deleted node are read (read step) and their value is compared (compare step). The element with smallest value is promoted and replaces the deleted element (write step). The same procedure repeats on each remaining level down the tree, selecting one children node to fill the empty node created by the last promotion. An example of a delete operation can be seen in figure \ref{fig:DeleteNaiveSHQ}. Read operations only consist of a locate phase.

\subsubsection{Insert operation}
\begin{figure}[ht!]
\centering
\includegraphics[width=6cm]{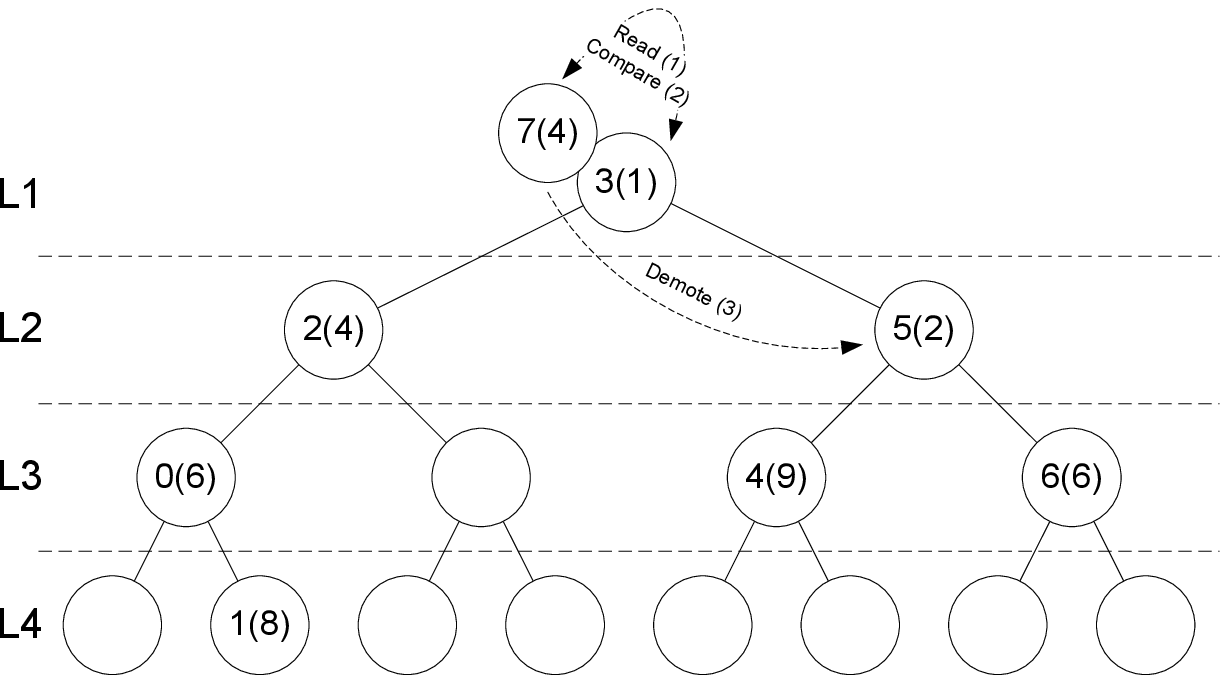}
\includegraphics[width=6cm]{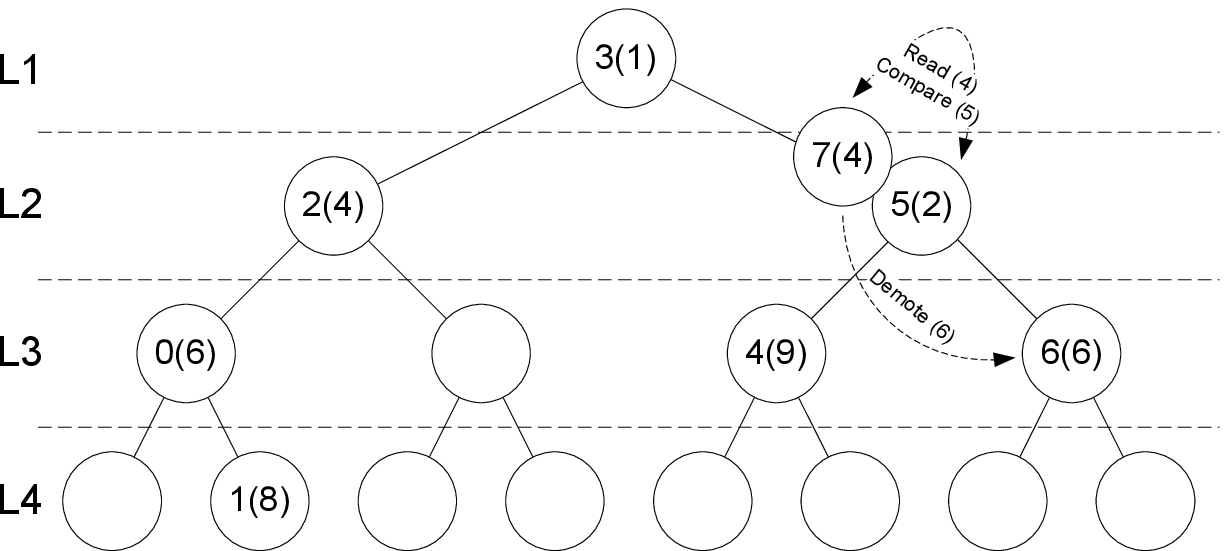}
\includegraphics[width=6cm]{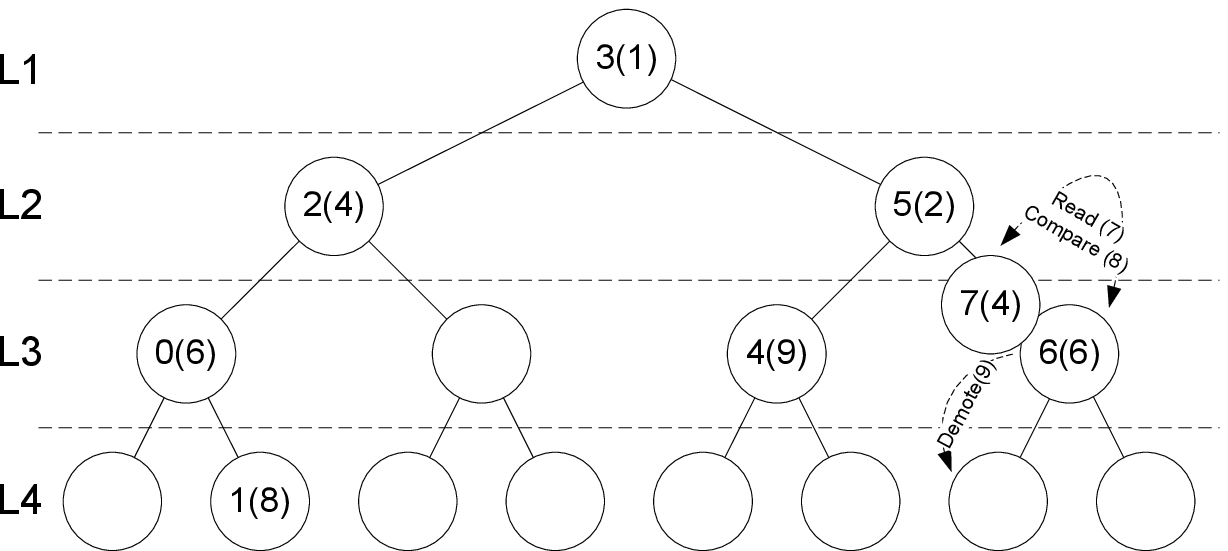}
\includegraphics[width=6cm]{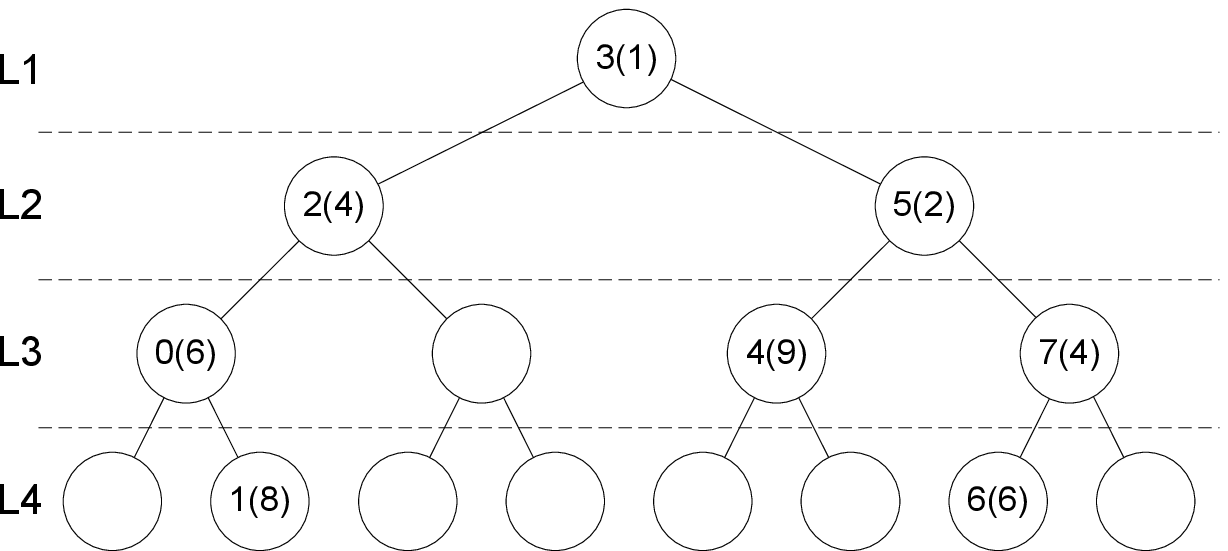}
\caption[Insert operation in the structured heap queue]{Insert operation in the structured heap queue. The pair of numbers in a node indicates the identification number (ID) and, inside parentheses, the value of the element. The operation consists in the insertion of the element \#7 with value 4. The operation's steps are indicated on the arrows along with their order in parentheses. \textbf{Top left} Initial state of the queue and first steps of the insert operation. The node containing element \#3 is read and its value is compared with that of element \#7. Element \#7 is demoted. \textbf{Top right} Intermediate steps of the insert operation. The node containing element \#5 is read and its value is compared with that of element \#7. Element \#7 is demoted. \textbf{Bottom left} Last steps of the delete-insert operation. The node containing element \#6 is read and its value is compared with that of element \#7. Element \#7 is written back in the node to replace element \#6 this one is demoted. \textbf{Bottom right} Final state of the queue after the insert operation.}
\label{fig:InsertNaiveSHQ}
\end{figure}

When an element is inserted, a certain number of elements have to be demoted to make room for it. On each level, a node is read (read step) and its value is compared with the inserted element's value (compare step). Whichever has the lowest value is written back in the node (write step) and the other one is demoted. This demoted element defines the path followed by the insert operation is it progresses down the tree. An example of an insert operation is shown in figure \ref{fig:InsertNaiveSHQ}.

\subsection{Design of the structured heap queue}
\begin{figure}[ht!]
\centering
\includegraphics[width=12cm]{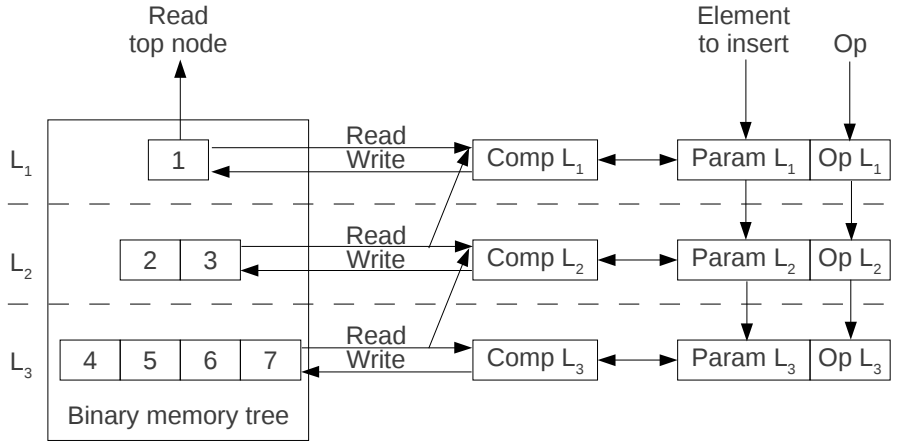}
\caption[Simple block diagram of the structured heap queue]{Simple block diagram of the structured heap queue. The same building block is attached to each level ``$\mathrm{L_{i}}$" of the binary memory tree. It consists of a read and a write port to level ``$\mathrm{L_{i}}$" of the memory tree, one read port to level ``$\mathrm{L_{i+1}}$", a comparator (``Comp") and two registers (``Param" and ``Op"). These registers hold the current operation type and parameters, for example the element to insert. This information is passed to the next level as the operation progresses down the tree. A separate port exists to read the information of the top node from outside the queue.}
\label{fig:BlockDiagSHQ}
\end{figure}

As in the PHQ \citep{IoannouKatevenis2007}, operations in the SHQ are pipelined. Each level of the tree is a pipeline stage. As soon as an operation is passed to a lower stage of the pipeline, another one can be serviced at the current level. The heap property is always satisfied and the lowest value element occupies the root node at all times, even if operations are still going on in lower levels of the tree. The stages of the pipeline are identical and consist of some hardware resources. As depicted in section \ref{sec:Mechanics}, all operations in the SHQ execute the same three steps: read, compare and write-back. Each stage of the pipeline thus has a read and a write port to its corresponding level of the memory tree and a 2-input comparator. Insert operations need to pass the element to insert down the tree, and a register is required for this purpose. Delete operations deal with data on two levels of the tree, 1 node and its 2 children nodes, and so require access to the read port from one level below. Figure \ref{fig:BlockDiagSHQ} show a simple block diagram of the SHQ.

\subsubsection{Consecutive operations}
\begin{figure}[ht!]
\centering
\includegraphics[width=6cm]{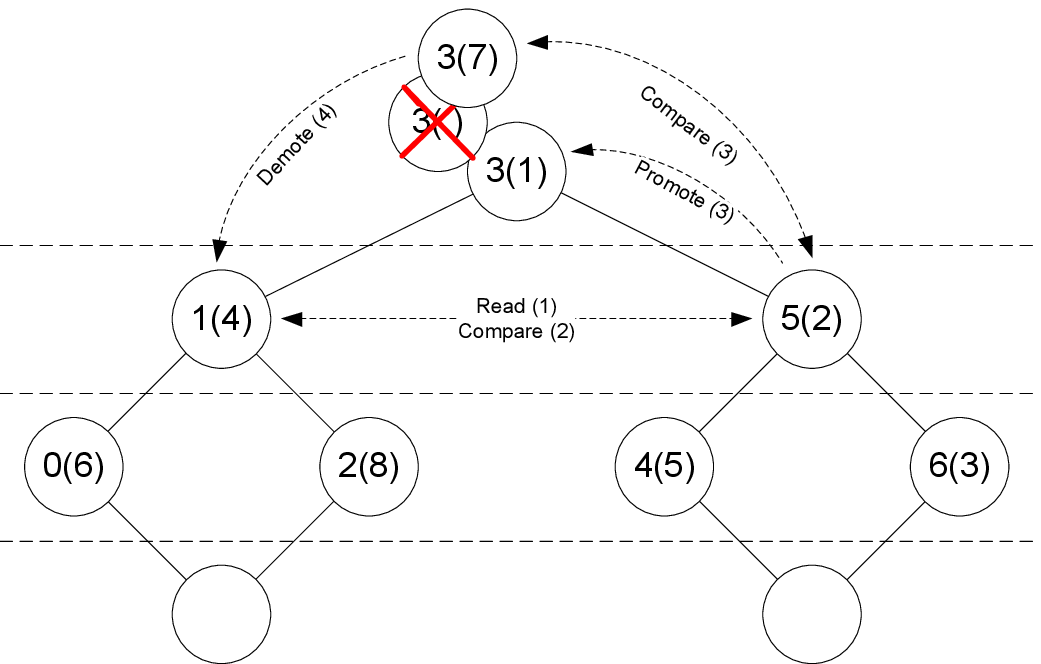}
\includegraphics[width=6cm]{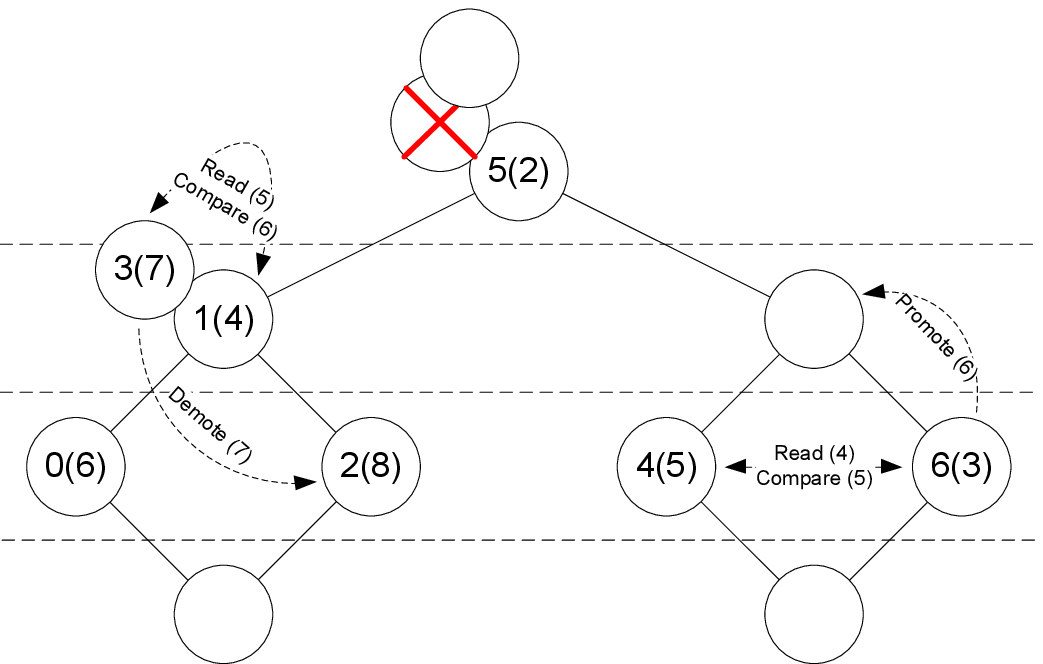}\\
\includegraphics[width=6cm]{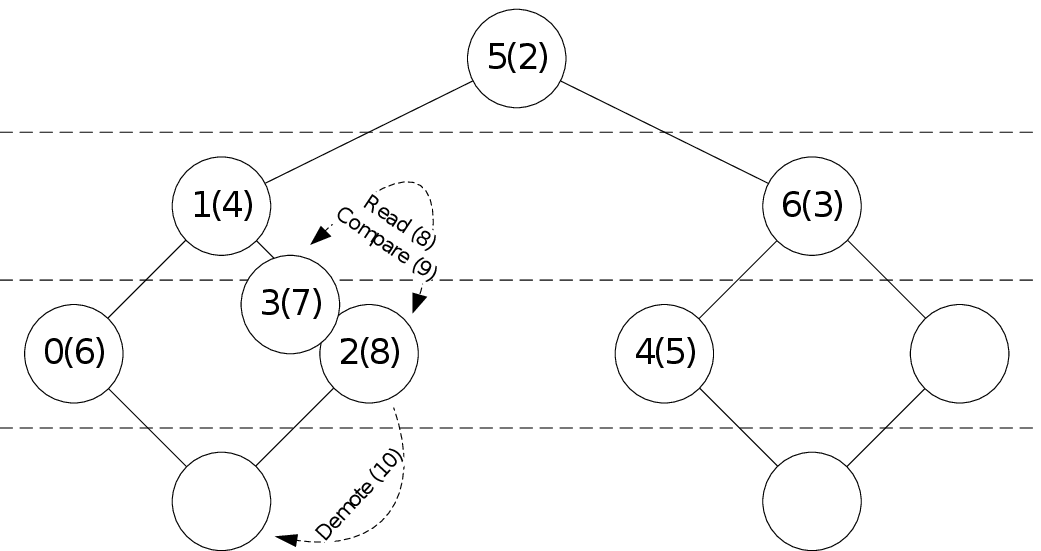}
\includegraphics[width=6cm]{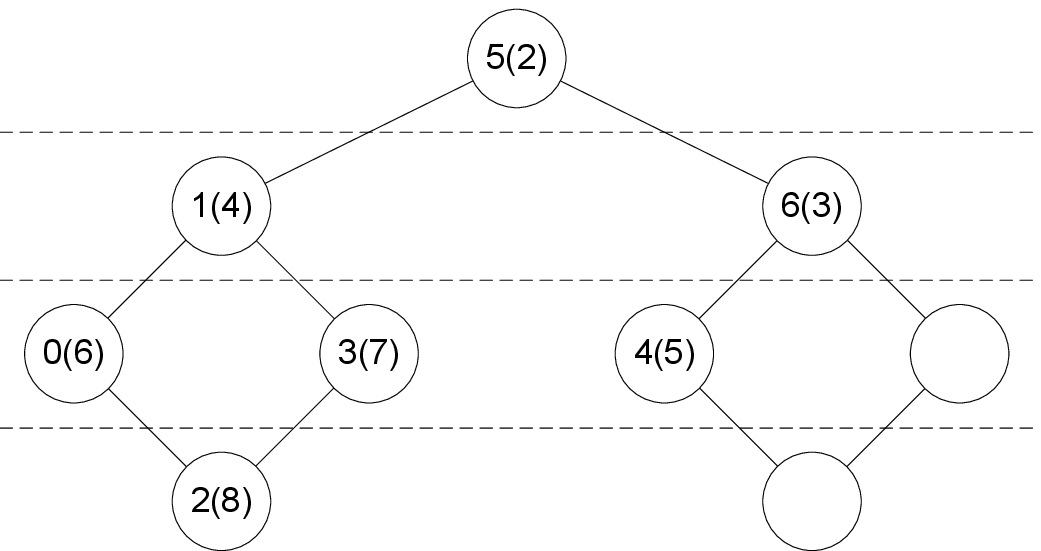}
\caption[Delete-insert operation in a memory-optimized structured heap queue]{Delete-insert operation in a memory-optimized structured heap queue. The pair of numbers in a node indicates the identification number (ID) and, inside parentheses, the value of the element. The operation consists in the deletion of the element \#3 directly followed with the insertion of the same element, changing its value. The operation's steps are indicated on the arrows along with their order in parentheses. \textbf{Top left} Initial state of the queue and first steps of the delete-insert operation. The nodes containing elements \#1 and \#5 are read and their value is compared. Element \#5 is promoted to replace element \#3. At the same time, the inserted element \#3 is compared with newly promoted element \#5 resulting in the demotion of element \#3. \textbf{Top right} Intermediate steps of the delete-insert operation. The nodes containing elements \#4 and \#6 are read and their value is compared. Element \#6 is promoted. At the same time, the node containing element \#1 is read and its value is compared with that of element \#3. Element \#3 is demoted. \textbf{Bottom left} Last steps of the delete-insert operation. The delete part of the operation is completed. The node containing element \#2 is read and its value is compared with that of element \#3. Element \#3 is written back in the node to replace element \#2 as this one is demoted. \textbf{Bottom right} Final state of the structured queue after the delete-insert operation.}
\label{fig:DelInsSHQ}
\end{figure}

The operations in the structured heap queue can be cascaded, just like in the PHQ. Figure \ref{fig:DetailOps} shows how several delete and insert operations can be issued back to back if reads, comparisons and write-backs each take one clock cycle to execute. Refer to figures \ref{fig:DeleteNaiveSHQ} and \ref{fig:InsertNaiveSHQ} for detailed description of the steps involved in the delete and insert operations. Delete operations take longer to execute than insert operations because they deal with data on two levels of the tree. Interleaved delete and insert operations are also shown in figure \ref{fig:DetailOps}. Figure \ref{fig:DelInsSHQ} shows an example of a delete-insert operation in a 4-level memory-optimized structured heap queue introduced in the next section.

\begin{figure}[ht!]
\centering
\begin{tabular}{l|cccccccc}
 & \multicolumn{8}{c}{\textbf{Clock cycle}} \\
\cline{2-9}
 & \textbf{1} & \textbf{2} & \textbf{3} &
\textbf{4} & \textbf{5} & \textbf{6} & \textbf{7} & \textbf{8} \\
\hline
{\small $1^{\mathrm{st}}$ Delete} & \footnotesize r$(\mathrm{L_{2}})_{\times2}$ & \footnotesize c$(\mathrm{L_{1}})$ & \footnotesize w$(\mathrm{L_{1}})$ & \footnotesize r($\mathrm{L_{3}})_{\times2}$ & \footnotesize c$(\mathrm{L_{2}})$ & \footnotesize w$(\mathrm{L_{2}})$ & \footnotesize r($\mathrm{L_{4}})_{\times2}$ & \footnotesize c$(\mathrm{L_{3}})$ \\
{\small $2^{\mathrm{nd}}$ Delete} & - & - & - & - & - & - & \footnotesize r$(\mathrm{L_{2}})_{\times2}$ & \footnotesize c$(\mathrm{L_{1}})$ \\
\hline
{\small $1^{\mathrm{st}}$ Insert} & \footnotesize r$(\mathrm{L_{1}})$ & \footnotesize c$(\mathrm{L_{1}})$ & \footnotesize w$(\mathrm{L_{1}})$ & \footnotesize r$(\mathrm{L_{2}})$ & \footnotesize c$(\mathrm{L_{2}})$ & \footnotesize w$(\mathrm{L_{2}})$ & \footnotesize r$(\mathrm{L_{3}})$ & \footnotesize c$(\mathrm{L_{3}})$ \\
{\small $2^{\mathrm{nd}}$ Insert} & - & - & - & \footnotesize r$(\mathrm{L_{1}})$ & \footnotesize c$(\mathrm{L_{1}})$ & \footnotesize w$(\mathrm{L_{1}})$ & \footnotesize r$(\mathrm{L_{2}}$) & \footnotesize c$(\mathrm{L_{2}})$ \\
\hline
{\small $1^{\mathrm{st}}$ Delete} & \footnotesize r$(\mathrm{L_{2}})_{\times2}$ & \footnotesize c$(\mathrm{L_{1}})$ & \footnotesize w$(\mathrm{L_{1}})$ & \footnotesize r$(\mathrm{L_{3}})_{\times2}$ & \footnotesize c$(\mathrm{L_{2}})$ & \footnotesize w$(\mathrm{L_{2}})$ & \footnotesize r$(\mathrm{L_{4}})_{\times2}$ & \footnotesize c$(\mathrm{L_{3}})$ \\
{\small $1^{\mathrm{st}}$ Insert} & - & \footnotesize r$(\mathrm{L_{1}})$ & \footnotesize c$(\mathrm{L_{1}})$ & \footnotesize w$(\mathrm{L_{1}})$ & \footnotesize r$(\mathrm{L_{2}})$ & \footnotesize c$(\mathrm{L_{2}})$ & \footnotesize w$(\mathrm{L_{2}})$ & \footnotesize r$(\mathrm{L_{3}})$ \\
{\small $2^{\mathrm{nd}}$ Delete} & - & - & - & - & - & - & - & \footnotesize r$(\mathrm{L_{2}})_{\times2}$ \\
\hline
\end{tabular}
\caption{Cascading delete operations (top), insert operations (center) and interleaved delete and insert operations (bottom) in the structured heap queue. Operations consist of a read (r), a compare (c) and a write (w) step. The ``$\mathrm{L_{i}}$" inside parentheses indicates the level of the tree where the read, compare or write takes place. The ``{\footnotesize $ _{\times 2}$}" subscript indicates that 2 simultaneous reads are performed (this happens in delete operations, as the information of two children nodes is read). A ``-" sign means the operation is not started yet. A certain delay must be met before a second operation can start to ensure it will not read data which might be overwritten by a previous operation (e.g. the first delete operation writes information of L2 during clock cycle 6, the second delete cannot be issued before clock cycle 7 because it needs to read of L2).}
\label{fig:DetailOps}
\end{figure}

\subsection{Memory-optimized structured heap queue} \label{sec:MemOpt}
Each element in a SHQ is assigned a unique path from the root node to one of the leaf nodes. Thus, there can be as many elements in the queue as there are nodes in the last layer of the tree. An $L$-level SHQ can store up to $2^{L-1}$ elements for $2^{L}$ nodes, as shown in figure \ref{fig:SchemNaiveSHQ}. In the PHQ, the tree is balanced, resulting in a more compact binary tree and lower memory requirement than in the SHQ. In comparison, an $L$-level PHQ can sort twice as many elements using the same amount of memory.

Assuming elements inserted in the queue are given unique IDs, the amount of memory required by the SHQ can be reduced. This is possible because the elements will distribute over the branches of the tree as they are inserted and will never fill the last level of the tree. A 3-level SHQ can store up to 4 elements and comprises 7 nodes. When an element is first inserted in this tree, it settles in the root node on level $\mathrm{L_{0}}$, as the rest of the tree is empty. A second inserted element will either push the first element down or will itself end up in level $\mathrm{L_{1}}$. Whatever the case, the root node and one node on level $\mathrm{L_{1}}$ will be occupied. Upon insertion of a $\mathrm{3^{rd}}$ element, two scenarios are possible. In scenario 1, the insertion causes an element to end up in the still free node of level $\mathrm{L_{1}}$. Before the insertion of the $\mathrm{4^{th}}$ element, the first 2 levels of the tree would be filled and the $\mathrm{3^{rd}}$ level would be empty. Upon insertion of the last element, one element will be pushed down in a leaf node, on level $\mathrm{L_{2}}$, and the other 3 leaf nodes will remain empty. In scenario 2, the $\mathrm{3^{rd}}$ insertion causes all three elements in the tree to lie on a single branch, with one element on each level of the tree. These 3 elements will be located in the same half of the tree, leaving the other half empty. Because element IDs are unique, the $\mathrm{4^{th}}$ insertion will necessarily result in one element being placed in the still empty node of level $\mathrm{L_{1}}$, again leaving 3 leaf nodes empty.

\begin{figure}[ht!]
\centering
\includegraphics{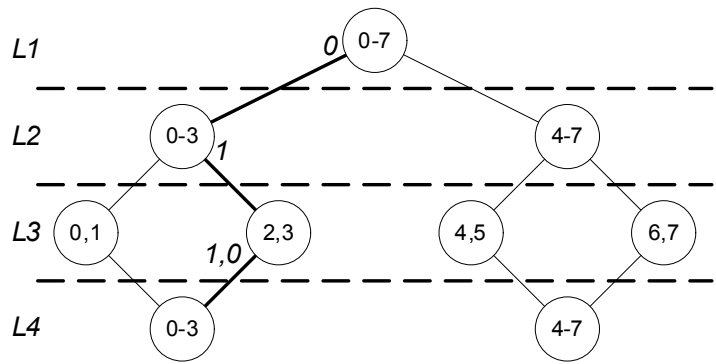}
\caption[Memory-optimized structured heap queue]{Memory-optimized structured heap queue. The numbers in a node indicate which elements can possibly occupy this node. In the memory-optimized SHQ, paths are shared by pairs of 2 elements and leaf nodes are shared by 4 elements. In the figure, the path shown in bold is shared by elements \#2 and \#3. This 4-level memory-optimized SHQ can handle 8 and contains 9 nodes. The last level of the tree uses 75\% less memory than in the naive implementation of the SHQ.}
\label{fig:MemOptSHQ}
\end{figure}

In a 3-level SHQ, only one leaf node can be occupied. Augmenting the 3-level queue to 4 levels is done by adding another 3-level queue next to the first one and a new root node on top of them. The 4-level queue can now sort up to 8 elements. Still, only one element can go all the way down each of the 3-level queues. That is, only two of the 8 available leaf nodes are actually useful. This reasoning holds true whatever the size of the queue. The memory-optimized SHQ reduces the size of the last level of the memory tree by 75\%. It can store $2^{L-1}$ elements using $1.25 \times 2^{L-1}$ nodes, which only represents a 25\% increase in memory over the PHQ. An example of a 4-level memory-optimized SHQ is shown in figure \ref{fig:MemOptSHQ}. The last level of a memory-optimized SHQ has to be implemented differently than the other ones, since each of its nodes has 2 parent nodes. Also, each path down the tree in a memory-optimized SHQ is shared by 2 elements.

\section{The Structured Heap Queue as an event queue}\label{sec:SNN}
In this section, we demonstrate the use of the SHQ in an event-driven SNN implemented on FPGA. The SNN is based on the Oscillatory Dynamic Link Matcher (ODLM) introduced by \citet{PichevarEtAl2006}. The hardware architecture is inspired by preliminary work from \citet{DHaene2010}. A very different approach has already been used to port the SNN to hardware \citep{CaronEtAl2011}, using a large array of 1-bit wide processing units and a time-driven strategy. This massively parallel implementation is efficient when several synchronized neurons spike in the same time step but slows down in periods of low activity. As it is a time-driven implementation, a comparison with the present work is out of the scope of this paper.

The ODLM uses the synchronization of spikes in a network of spiking neurons to accomplish binding \citep{Milner1974,VonDerMalsburg1981}. This allows the ODLM to perform different signal processing tasks such as image segmentation, image matching \citep{PichevarEtAl2006} and sound source separation \citep{PichevarRouat2007}. In the following section, we describe the hardware implementation of the ODLM, present the resource usage and performance of the design and show the results of an image segmentation experiment.

\subsection{The Oscillatory Dynamic Link Matcher}
To facilitate the port to hardware, some of the features of the original ODLM were not implemented on the FPGA. These features, namely global inhibition and a dynamic normalization of synaptic weights, are not essential to the basic operation of the system. The implemented model and the method used to accomplish image segmentation on hardware are described here.

\subsubsection{The neuron model}
A Leaky Integrate and Fire (LIF) neuron model \citep{GerstnerKistler2002} is used to approximate the behavior of relaxation oscillators. The membrane potential of an isolated neuron with initial potential zero follows 
\begin{equation}
\label{eq:oLIF}
	p_{i}(t) = \frac{I_{0}}{\tau}\left(1-e^{-t / \tau}\right),
\end{equation}
where $p_{i}(t)$ is the membrane potential of neuron $i$ at time $t$, and $I_{0}$ (input current) and $\tau$ (membrane time constant) are parameters. When the membrane potential of a neuron reaches the threshold value $p^{\theta}$, it is reset and a spike is emitted. A neuron is reset by subtracting the value $p^{\theta}$ from its membrane potential. A neuron defined by equation (\ref{eq:oLIF}) fires periodically, if $p^{\theta}$ is smaller than $\frac{I_{0}}{\tau}$, and acts like a relaxation oscillator.

When the membrane potential of neuron $i$ reaches the threshold, the neuron emits a spike. This spike will affect all post-synaptic neurons $j$ to which it is connected. The neurons $j$ will have their membrane potential instantly increased by an amount $w_{ij}$. This is shown in equation (\ref{eq:Spike}), where $t_{s_{i}}$ is the time at which neuron $i$ spikes and $w_{ij}$ is the synaptic weight between pre-synaptic neuron $i$ and post-synaptic neuron $j$.
\begin{equation}
\label{eq:Spike}
	p_{j}(t_{s_{i}}) \leftarrow p_{j}(t_{s_{i}}) + w_{ij}
\end{equation}

Synapses are defined by a positive scalar weight computed using equation (\ref{eq:Weights}), where $w_{ij}$ is the synaptic weight connecting neuron $i$ to neuron $j$, $w^{MAX}$ is the maximum value for a weight, $f_{i}$ is a feature of the input associated to neuron $i$, and $\alpha$ and $\delta$ are parameters which must be adapted to the processing task.
\begin{equation} \label{eq:Weights}
	w_{ij} = w^{MAX} \times \left(1 - \frac{1}{1 + e^{-\alpha\left(|f_{i}-f_{j}| + \delta\right)}}\right)
\end{equation}

\subsubsection{Image segmentation using the ODLM}
The topology of the network and the features used depend on the signal processing task to accomplish. For image segmentation, a flat, 2-dimensional nearest neighbor network, where each neuron is bidirectionally connected to its 8 neighbors, is used. In this work, each pixel of the image is associated to a neuron in the network and the gray level of the image's pixels (an integer number in the range 0 to 255) is used as the feature to compute synaptic weights. The segmentation of a $N$-pixel image thus requires a network of $N$ neurons and $8\times N$ synaptic connections. Once the weight values are calculated, the membrane potential of the neurons is initialized randomly and the network is run until convergence is reached. As they interact, neurons connected by strong synaptic weights will tend to synchronize, firing at the same instant. When the groups of synchronized neurons remain unchanged for a number of neuron oscillations, the network is said to have reached convergence. The final membrane potential value of the neurons is analyzed to determine the image's segmentation: neurons with similar final membrane potential values are part of the same segment.

\subsection{Event-driven ODLM}\label{sec:Processing}
According to equation (\ref{eq:oLIF}), the potential of an isolated neuron increases continuously. Inverting this equation yields equation (\ref{eq:InvoLIF}), where $t^{p_{0}\rightarrow p}$ is the time it takes for a neuron with initial potential 0 to reach potential $p$.

\begin{equation} \label{eq:InvoLIF}
	t^{p_{0}\rightarrow p}(p) = -\tau \ln\left(1-p\frac{\tau}{I_{0}}\right)
\end{equation}
Using equation (\ref{eq:InvoLIF}), it is possible to calculate a neuron's next firing time $$t_{s} = t + t^{p(t)\rightarrow p^{\theta}} = t + t^{p_{0}\rightarrow p^{\theta}} - t^{p_{0}\rightarrow p(t)}.$$

The state of the neurons is stored in memory as the predicted time of their next spike. Processing an event involves modifying neuron potentials, and not directly their predicted firing times. A neuron's predicted firing time must be translated into a membrane potential before equation (\ref{eq:Weights}) is applied. A new membrane potential is obtained which is translated back into a new predicted firing time and written in memory. Equations (\ref{eq:oLIF}) and (\ref{eq:InvoLIF}) define the mapping between firing time and membrane potential. Each time the SNN processes an event, the current time $t$ is updated to this event's time of occurrence $t_{s_{i}}$.

When a spike is processed, the pre-synaptic neuron is reset and a synaptic weight is added to the membrane potential of each of the post-synaptic neurons. These two operations are very similar, as they both consist in the modification of the membrane potential of a neuron. Algorithm \ref{alg:Event} is used to apply the effect of an incoming spike on a post-synaptic neuron, that is, to add a synaptic weight to its membrane potential. 

\algsetup{indent=2em}
\begin{algorithm}[ht!]
\caption{Processing of a spike}\label{alg:Event}
\begin{algorithmic}[1]
\medskip
\STATE Identify the post-synaptic neuron to process\label{algline:Topo}
\STATE Retrieve the neuron's firing time and pixel value\label{algline:Memory}
\STATE Calculate the neuron's current membrane potential\label{algline:Model}
\STATE Calculate the synaptic weight\label{algline:Weight}
\STATE Calculate the neuron's new membrane potential\label{algline:Synapse}
\STATE Calculate the neuron's new next firing time\label{algline:Inverse}
\STATE Update the neuron's information in memory\label{algline:WriteB}
\end{algorithmic}
\end{algorithm}

The same algorithm can be used to reset the pre-synaptic neuron, with a single difference: instead of adding a synaptic weight to the membrane potential of the neuron, step \ref{algline:Synapse} of algorithm \ref{alg:Event} resets the membrane potential (and discards the synaptic weight calculated in step \ref{algline:Weight}). For the processing of an event, algorithm \ref{alg:Event} is executed $N+1$ times, where $N$ is the number of post-synaptic neurons. During the extra execution, the pre-synaptic neuron is reset.

\subsection{Hardware implementation} \label{sec:Implementation}
\begin{figure}[ht!]
\centering
\includegraphics{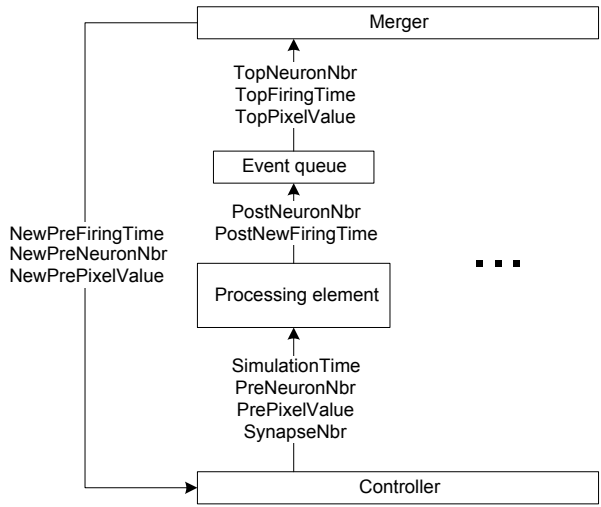}
\caption{\textbf{A high level view of the HSNN}. The controller receives the next event's information from the merger and forwards it to the processing element (PE) along with some control signals. The PE identifies the post-synaptic neurons, and computes the required synaptic weights and the new state of the processed neurons. The new state of the post-synaptic neurons is sent to the event queue for update and the queue is sorted on-the-fly. The merger reads the event on top of the event queue and sends it to the controller. If several PEs and event queues are used, the merger chooses the very next event to happen and the controller distributes the processing load among the PEs.}
\label{fig:EDNeuroSpike}
\end{figure}

The SNN is an implementation of algorithms \ref{alg:ED} and \ref{alg:Event} on a FPGA. It consists of a controller, a processing element (PE), an event queue and a merger. The PE implements algorithm \ref{alg:Event} and can be duplicated to reduce processing time. Figure \ref{fig:EDNeuroSpike} gives an overview of the resulting system and its various components are detailed in the following sections.

\subsubsection{Controller}
The controller realizes step \ref{algline:EDtime} of algorithm \ref{alg:ED}. It receives from the merger the next event to be processed, then updates the simulation time and forwards the event for processing. If several PEs are implemented, the post-synaptic neurons to be processed can be evenly distributed among them. In the present work, all neurons are serially processed by a single PE, but several PEs could very well be used. The controller also takes care of the communications with the computer host.

\subsubsection{Processing element}
\begin{figure}[ht!]
\centering
\includegraphics[width=12cm]{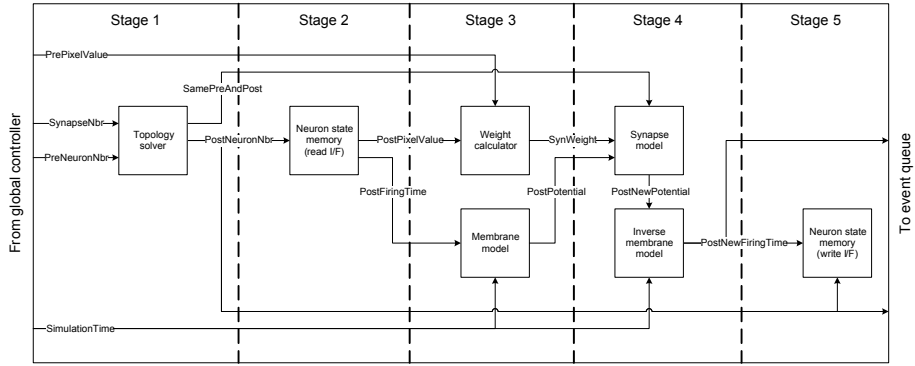}
\caption{\textbf{A processing element} of the hardware event-driven system. The inputs to the pipeline are coming from the controller and the outputs are sent to the event queue. The topology solver identifies the post-synaptic neurons associated with the event. The neuron state memory stores the state of all neurons. It has one read and one write interface, the latter being used for write-back in the last stage of the pipeline. The weight calculator computes a synaptic weight based on the pixel value of the processed neurons. The membrane model translates the firing time of a neuron into a membrane potential value. The synapse model either adds a synaptic weight to the membrane potential of a neuron or resets this neuron. The inverse membrane model translates the new membrane model back into a firing time. Signals crossing dashed lines are delayed appropriately (delay elements not shown).}
\label{fig:ProcessingElement}
\end{figure}

A PE instantiates step \ref{algline:EDtime} of algorithm \ref{alg:ED} as detailed in algorithm \ref{alg:Event}. An overview of a PE is given in figure \ref{fig:ProcessingElement}. PEs receive a synapse number, a pre-synaptic neuron ID, a weight parameter and the simulation time from the controller. A 5-stage pipeline processes the neurons involved in the event, computing their new state. In the first stage of the pipeline, step \ref{algline:Topo} of algorithm \ref{alg:Event} is executed. Stage 2 executes the $\textrm{\ref{algline:Memory}}^{\mathrm{nd}}$ step of the algorithm. Steps \ref{algline:Model} and \ref{algline:Weight} are executed in the $\textrm{3}^{\mathrm{rd}}$ stage of the pipeline. The $\textrm{4}^{\mathrm{th}}$ stage implements steps \ref{algline:Synapse} and \ref{algline:Inverse} and the last stage executes step \ref{algline:WriteB} . Each stage of the pipeline takes one clock cycle to execute. Most of the computations are performed by memory look-up. To implement an equation by a look-up table\footnote{In the paper, the term look-up table can refer to two different things. For clarity, ``look-up table" will designate the implementation of an equation by memory look-up while the acronym LUT will be used for the FPGA resource (see section \ref{sec:Resource})}, one has to pre-compute the output of the equation for several (or all) input values and save the result in memory. To evaluate the equation, the input value for which the equation has to be computed is used as the address of the memory to retrieve the output value of the equation. In the processing pipeline, this technique is used for the weight calculation, the neuron model and the inverse neuron model. A description of all the stages of the pipeline follows.

\begin{figure}[ht!]
\centering
\includegraphics[width=8.6cm]{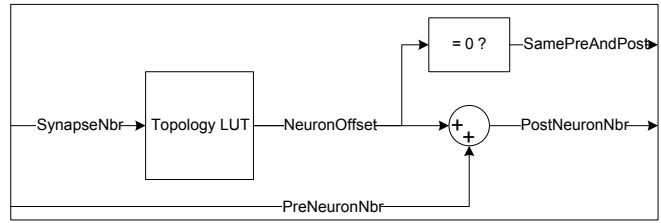}
\caption{\textbf{The topology solver} is the $1^{\mathrm{st}}$ stage of the pipeline in a processing element. The synapse number is used as the address to the topology look-up table which outputs an offset to add to the pre-synaptic neuron ID. If the offset is 0, a flag is set, indicating that the pre-synaptic and post-synaptic neuron IDs are equal.}
\label{fig:TopologySolver}
\end{figure}

\paragraph{Topology solver}
The topology solver identifies the neurons involved in the current event and outputs their ID. The block outputs the neuron IDs one at a time and indicates with a flag if the current ID corresponds to the pre-synaptic neuron or to one of the post-synaptic neurons (see section \ref{sec:Processing}). A functional view of the topology solver is shown in figure \ref{fig:TopologySolver}.

\paragraph{Neuron state memory}
In the second stage of the pipeline, the post-synaptic neuron's information is retrieved from the neuron state memory. This memory stores the state variables and parameters of the neurons. For the image segmentation experiment, a firing time and a pixel value are stored for each neuron. Neuron IDs are used to address the neuron state memory.

\begin{figure}[ht]
\centering
\includegraphics[width=8.6cm]{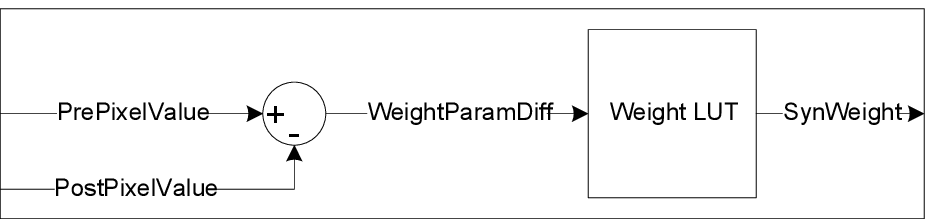}
\caption{\textbf{The weight calculator} composes, with the membrane model, the $3^{\mathrm{rd}}$ stage of the pipeline of a processing element. The pixel difference is used as the address to a look-up table which outputs the synaptic weight.}
\label{fig:WeightCalc}
\end{figure}

\paragraph{Weight calculator}
The weight calculator takes two pixel values and computes the corresponding synaptic weight. The calculation of the weight depends on the difference between the pre-synaptic neuron's pixel value and the post-synaptic neuron's pixel value. A functional view of the weight calculator is shown in figure \ref{fig:WeightCalc}.

\begin{figure}[ht]
\centering
\includegraphics[width=8.6cm]{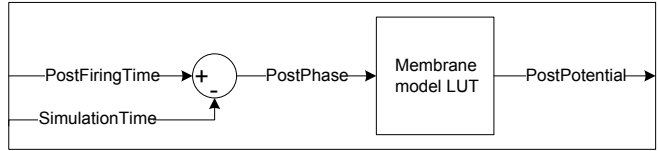}
\caption{\textbf{The membrane model}, along with the weight calculator, is part of the $3^{\mathrm{rd}}$ pipeline stage in a processing element. The subtraction of the simulation time to the firing time of a neuron results in a value which represents how far from the threshold the neuron is. This value is used as the address of a look-up table to get the membrane potential of the neuron.}
\label{fig:MembraneModel}
\end{figure}

\paragraph{Membrane model}
The membrane model calculates the membrane potential at the current simulation time given a firing time. A functional view of the membrane model is shown in figure \ref{fig:MembraneModel}.

\begin{figure}[ht]
\centering
\includegraphics[width=8.6cm]{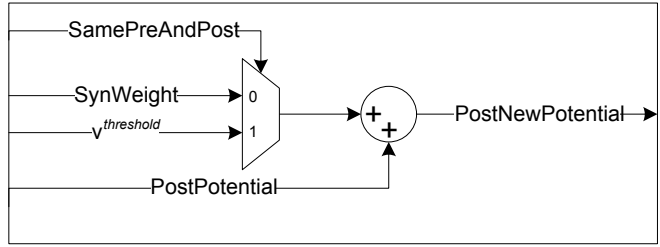}
\caption{\textbf{The synapse model} is combined with the inverse membrane model to form the $4^{\mathrm{th}}$ stage of the pipeline in a processing element. The post-synaptic neurons' membrane potential is added to the synaptic weight if the input flag is 0. Otherwise, the value $p^{\theta}$ is subtracted from the potential to reset the pre-synaptic neuron.}
\label{fig:SynapseModel}
\end{figure}

\paragraph{Synapse model}
The synapse model has a different role depending on the value of the flag outputted by the topology solver. If the flag is 0, then the post-synaptic neuron's potential is added to the synaptic weight. Otherwise, the processed neuron is the pre-synaptic neuron, and its potential is reset. A functional view of the synapse model is shown in figure \ref{fig:SynapseModel}.

\begin{figure}[ht]
\centering
\includegraphics[width=8.6cm]{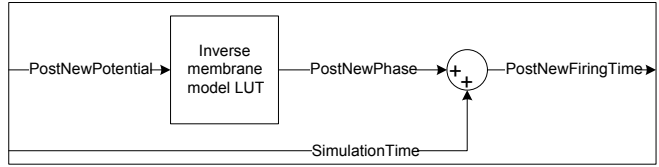}
\caption{\textbf{The inverse membrane model} and the synapse model form the $4^{\mathrm{th}}$ stage of the pipeline. The post-synaptic neuron's membrane potential is the address to a look-up table which outputs the amount of time left until the neuron fires. This value is added to the simulation time to get the new neuron's firing time.}
\label{fig:InverseMembraneModel}
\end{figure}

\paragraph{Inverse membrane model}
The inverse membrane model computes the new firing time of a neuron based on its new membrane potential. A functional view of the inverse membrane model is shown in figure \ref{fig:InverseMembraneModel}.

\paragraph{State memory write back}
Once the new firing time is calculated, it is written back into the neuron state memory. This is done in the last stage of the processing pipeline.

\subsubsection{Event queue}\label{sec:EventQueue}
The event queue is implemented by the structured heap queue described in section \ref{sec:SHQ}. The interface between the PE and the event queue is very simple. At initialization, insert operations are issued to populate the event queue with the neurons' initial predicted firing time. During the processing, when the PE computes the new firing time of a neuron, the information in the event queue must be updated. This is done by issuing a delete-insert operation with the neuron ID and the new firing time. For every neuron involved in an event, a delete-insert operation must be serviced by the event queue. The root node is then read to start the processing of the next event.

\subsubsection{Merger}
The merger is required when several PEs are used in parallel. It scans the top nodes of all the event queues to determine the very next event to be processed. In the present work, the merger simply passes the information of the element in the root node of the single event queue to the controller.

\section{Results} \label{sec:EDNS_Results}
A network of 65~536 neurons is implemented on a Xilinx XC5VSX50T FPGA clocked at 100 MHz. For the image segmentation experiment, each neuron has 8 synapses for a total of 524~288 implemented synapses. The next sections summarize the resource usage of the design and the result of the image segmentation experiment.

\subsection{Resource usage} \label{sec:Resource}
The membrane potential and firing time values are 13-bit wide so the membrane model and inverse membrane model look-up tables can each be implemented an assembly of 3 BRAMs. The pixel values and synaptic weights are respectively 8-bit and 9-bit wide. The resources usage in terms of flip-flops (FFs), look-up tables (LUTs) and block RAMs (BRAMs) for the whole system is given in table \ref{tab:EDNS_ResourcesWhole}, and table \ref{tab:EDNS_ResourcesByFunc} details these numbers for each functional block of the design. Please note that in a Xilinx FPGA, a LUT is a resource mainly used to implement logical operations. In the other sections of this paper, the term look-up table designates the implementation of an equation by memory look-up.

\begin{table}[t]
\caption{Resource usage for the whole system}
\label{tab:EDNS_ResourcesWhole}
\begin{center}
\begin{tabular}{lrrr}
\textbf{RESOURCE} & \textbf{USED} & \textbf{AVAILABLE} & \textbf{\% USED} \\
\hline
FFs      & 3 368   & 32 640   & 10\%\\
LUTs     & 4 673   & 32 640   & 14\%\\
Slices   & 2 855   & 8 160    & 35\%\\
BRAMs    & 130     & 132      & 98\%\\
\end{tabular}
\end{center}
\end{table}

\begin{table}[t]
\caption{Resource usage for each functional block}
\label{tab:EDNS_ResourcesByFunc}
\begin{center}
\begin{tabular}{lrrr}
\textbf{FUNCTION} & \textbf{FFs} & \textbf{LUTs} & \textbf{BRAMs} \\
\hline
Controller                          & 503   & 635   & 0 \\
Processing element                  & 19    & 73    & 50 \\
\hspace{1mm} Topology solver        & 16    & 26    & 0 \\
\hspace{1mm} Neuron memory          & 1     & 1     & 44 \\
\hspace{1mm} Weight calculator      & 0     & 8     & 0 \\
\hspace{1mm} Membrane model         & 1     & 14    & 3 \\
\hspace{1mm} Synapse model          & 0     & 10    & 0 \\
\hspace{1mm} Inverse membrane model & 1     & 14    & 3 \\
Event queue                         & 2 846 & 3 965 & 80 \\
Merger                              & N/A   & N/A   & N/A \\
\end{tabular}
\end{center}
\end{table}

\subsection{Image segmentation}\label{sec:Segmentation}
\begin{figure}[t]
\centering
\includegraphics[width=3.5in]{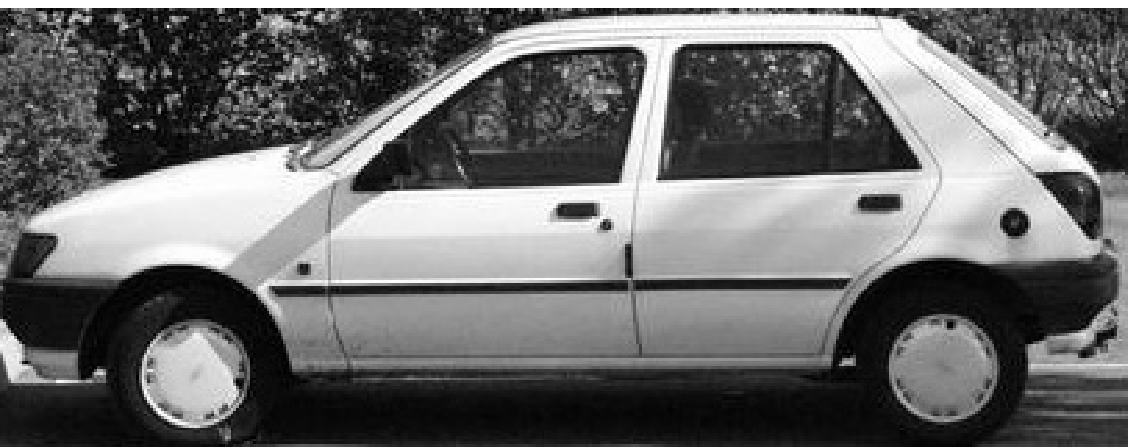}
\includegraphics[width=3.5in]{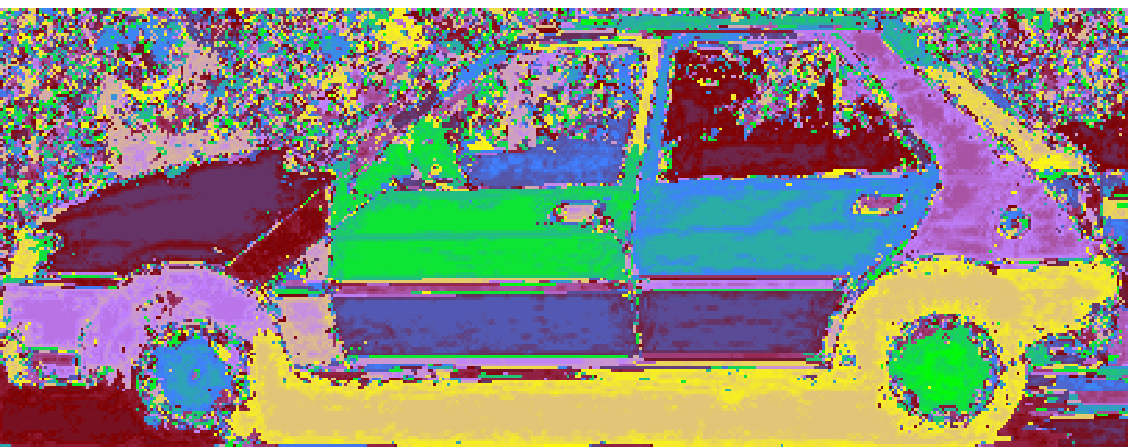}
\caption[Image segmentation experiment using the event-driven system]{Image segmentation experiment. \textbf{Top} The original 406$\times$158 pixel image. \textbf{Bottom} The segmented image, where different colors represent different segments. The segmentation is solely based on pixel values and the HSNN merges the lower part of the image with the tires of the car. Also, the foliage in the background is segmented with preservation of the texture.
}
\label{fig:EDNS_Seg}
\end{figure}

The results of an image segmentation task using the event-driven SNN are presented in this section. For this experiment, a host computer transfers the pixel values of the image and random initial potential of the neurons into the SNN. The host computer also populates the various look-up tables in the PE as well as the event queue. The SNN is then run for a given time and the final membrane potential values are sent back to the computer for visualization. The parameters of equations (\ref{eq:oLIF}) and (\ref{eq:Weights}), using the pixels' level of gray as features, are set to: $I_{0}=6.918$, $\tau=0.1447$, $v^{\theta}=1$, $w_{max}=0.0325$, $\alpha=100$, $\delta=6$. The network was run for 200 ms. The original and segmented images are presented in figure \ref{fig:EDNS_Seg}. The same task on a general purpose computer (Intel Core i5 @2.4GHz, 3GB RAM) executes in 1.9 seconds.

\section{Discussion}\label{sec:Discussion}
An event-driven SNN using the SHQ presents three major advantages: scalability, versatility and powerful resource sharing. These aspects are covered in the next sections. The main drawback of the SHQ is the memory overhead of 25\% compared to a PHQ and to \citeauthor{AgisEtAl2007}'s search algorithm. This overhead only is a constant scaling factor and the SHQ has the same $O(N)$ complexity in memory as the two other algorithms.

\subsection{Scalability}\label{sec:Complexity}
The scalability of a SNN defines the impact of increasing the number of neurons on performance and hardware resources used. In an event-driven SNN, this is greatly affected by the complexity of the event queue. Table \ref{tab:sortComp} compares the complexity in time, logic resources and memory of the SHQ and the pipelined search of \citet{AgisEtAl2007}. Operations on ``top element" refer to the next event to happen.

\begin{table}[ht]
\caption{Comparison of the complexity in logic, memory and time of the structured heap queue (SHQ) and \citeauthor{AgisEtAl2007}'s search algorithm, with $N$ the number of elements in the queue}
\label{tab:sortComp}
\begin{center}
\begin{tabular}{lrr}
 & \multicolumn{1}{c}{\textbf{SHQ}} & \multicolumn{1}{c}{\textbf{\citet{AgisEtAl2007}}} \\
\hline
Complexity in logic             & $O(\log(N))$ & $O(N)$* \\
Complexity in memory            & $O(N)$       & $O(N)$ \\
Complexity in time for          &              & \\
\hspace{1mm} Delete top element & $O(1)$       & $O(\log(N))$* \\
\hspace{1mm} Delete any element & $O(1)$       & $O(1)$ \\
\hspace{1mm} Insert             & $O(1)$       & $O(1)$ \\
\hspace{1mm} Read top element   & $O(1)$       & $O(\log(N))$* \\
\hspace{1mm} Read any element   & $O(\log(N))$ & $O(1)$ \\
\hline
\multicolumn{3}{p{10cm}}{*In \citep{AgisEtAl2007}, logic can be traded for time, and conversely, time for logic.}
\end{tabular}
\end{center}
\end{table}

The most important values are the complexity in logic, and the complexity in time of the delete top element and the read operations. \citeauthor{AgisEtAl2007}'s pipelined search uses a 4-level comparator tree. With this structure, it is possible to trade logic resources for time, by dividing the set of elements into subsets and processing each one serially. For example, $N$ can be doubled without increasing the amount of logic resources, but the search would take twice the amount of time to execute. If the processing time is kept constant, the number of comparators increases linearly with $N$. The SHQ, on the other hand, uses a binary memory tree and requires one 2-input comparator per level of the tree. This results in a much nicer logarithmic scaling of logic resources with the number of elements.

To delete or read the top element when using the pipelined search, the smallest value element must first be identified. This requires logarithmic time, with a radix depending on the exact structure of the comparator tree. With the SHQ, deleting the top element is done by reading the root node and issuing a delete operation with the correct ID. This operation requires a certain amount of time to complete, but it is not necessary to wait for it to be done before the queue is usable again. When the delete operation executes on the top level of the tree, it selects one of the children node and promotes it. The promoted node, readable in the root node as soon as the delete operation is passed down to the second level of the queue, is the new smallest value element of the queue. All operations, except for the read any element, effectively execute in constant time with respect to the number of elements. The read top element operation executes in 1 clock cycle, as a special read port is provided for this purpose (see figure \ref{fig:BlockDiagSHQ}).

The pipelined search has a better complexity than the SHQ for the read any element operation. Because they use an unsorted list, elements can be read very simply. This is not the case with the SHQ, in which a read operation might have to scan all the levels of the binary memory tree before it finds the element it is looking for. This logarithmic complexity does not compare very well with the constant time of the pipelined search. Nevertheless, this is not of much importance for the implementation of an event queue. As explained in section \ref{sec:Intro}, the read any node operation is not required in a SNN.

\begin{figure}[ht]
\centering
\includegraphics[width=8.6cm]{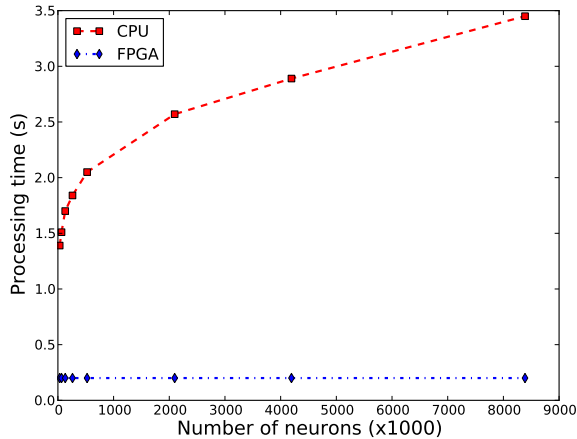}
\caption{Comparison of the processing speed for a software and a FPGA implementation of the SNN. For CPU, results show the amount of time required to process 250~000 spikes with networks of various sizes. The values reported for FPGA are based on the theoretical complexity given in section \ref{sec:Complexity}.}
\label{fig:ScalingCPU_FPGA}
\end{figure}

To show the benefits of the digital hardware SNN over an implementation on CPU, experiments were done with a software version of the SNN described in this paper. A software SNN of increasing size was run and the time required to process 250~000 spikes is reported in figure \ref{fig:ScalingCPU_FPGA}. The figure shows the logarithmic scaling of processing time with the number of neurons of the software implementation. To compare these results, the theoretical constant time scaling of the FPGA implementation is also shown with the real performance of the 65~536 neurons SNN as a baseline. Real experiments could not be run on FPGA because of the network size involved. With smaller networks, the logarithmic scaling of the software implementation is much less obvious.

\subsection{Resource sharing and performance}
The SNN implementation shown in this paper illustrates the power of resource sharing. A single PE is used, but a 65~536-neuron, 513~184-synapse SNN is implemented. This is typical of digital hardware circuits: resource sharing, or time multiplexing, allows great flexibility with regard to the amount of resources used. The implementation can be tailored to the specific needs of the designer and to the resources at his disposal. If resource shortage limits the design to only one PE, a SNN of arbitrary size can still be designed. If performance is an issue, spare resources can be used to duplicate the PE and cut the processing time. In the presented SNN, a PE can process one post-synaptic neuron every 7 clock cycles. It would be possible to modify the system we presented and use 9 PEs (the number of neurons involved in an event) with minimal added design effort. Each PE would process one of the involved neurons, resulting in a processing speed of one event per 7 clock cycles.

\subsection{Versatility}
As a last point, the SHQ can be used in virtually any digital hardware event-driven SNN. In this paper, we chose to use LIF neurons, a regular network topology, look-up tables computations, and so on. The SHQ is oblivious to all of this. Irregular network topologies or dynamic synaptic weights could be used. It does not matter if event times are exact or approximated. The SHQ only brings constraints on timing: insert operations can only be serviced every 3 clock cycles and delete-insert operations every 7 clock cycles.

\section{Conclusion}\label{sec:Conclusion}
In this paper, the structured heap queue (SHQ), was introduced as a good candidate to implement the event queue of an event-driven digital hardware spiking neural network. The use of the SHQ was demonstrated in the FPGA implementation of a SNN and tested with an image segmentation task. With the SHQ, it is possible to process one event every 7 clock cycles, no matter the size of the SNN. The SHQ is very similar to the pipelined heap queue \citep{IoannouKatevenis2007,BhagwanLin2000}, but it is especially well suited for the application. In the SHQ any existing element can be deleted in constant time, a crucial feature for the implementation of an event queue. An alternate solution to the SHQ is the pipelined search algorithm used in \citep{AgisEtAl2007}. The SHQ has a better complexity both in logic resources and in time for the operations required in an event queue. With the SHQ, doubling the number of events to manage results in double the amount of memory and one more 2-input comparator. Processing time is not affected. The SHQ can be put to use in virtually any event-driven digital hardware SNN.

\section*{Acknowledgements}
This work has been funded by the Natural Sciences and Engineering Research Council of Canada (NSERC) and the Fonds qu\'{e}b\'{e}cois de la recherche sur la nature et les technologies (FQRNT). The collaborative work between Université de Sherbrooke and Ghent University was made possible by a grant from NSERC's Michael Smith Foreign Study Supplements Program. It allowed for Sherbrooke's signal processing SNN ODLM \citep{PichevarEtAl2006} to be implemented using Ghent's event-driven hardware algorithms \citep{DHaene2010}.

\bibliographystyle{plain}
\bibliography{Bibliography_LCCaron}

\end{document}